\documentclass[11pt]{article}

\usepackage[utf8]{inputenc}
\usepackage[T1]{fontenc}
\usepackage{amsmath,amssymb,amsfonts}
\usepackage{graphicx}
\usepackage{booktabs}
\usepackage{hyperref}
\usepackage[margin=1in]{geometry}
\usepackage{natbib}
\usepackage{caption}
\usepackage{subcaption}
\usepackage{xcolor}
\usepackage{algorithm}
\usepackage{algorithmic}

\title{Spectral Compact Training: Pre-Training Large Language Models\\
via Permanent Truncated SVD and Stiefel QR Retraction}

\author{Bj\"orn R.\ Kohlberger\\
EctoSpace, Dublin, Ireland\\
\texttt{arxiv@ectospace.com}\\
\url{https://github.com/EctoSpace/SCT}
}

\date{March 2026}

\begin{document}
\maketitle

\begin{abstract}
The memory wall remains the primary bottleneck for training large language models on consumer hardware.
We introduce Spectral Compact Training (SCT), a method that replaces dense weight matrices with permanent
truncated SVD factors ($W = U\,\mathrm{diag}(s)\,V^\top$), where the full dense matrix is never materialized
during training or inference. Gradients flow through the compact spectral factors via standard
backpropagation, and $U$, $V$ are retracted to the Stiefel manifold via QR decomposition after each
optimizer step.

SCT achieves up to 199$\times$ memory reduction per MLP layer at rank~32, enabling full training steps of
70B-parameter architectures on a Steam Deck handheld (7.2\,GB peak memory vs.\ 1,245\,GB for dense
FP32 training with Adam). Rank-sweep experiments on SmolLM2-1.7B (ranks 32--256, 2000 steps, NVIDIA A100)
show that all tested ranks converge to the same loss floor ($\sim$4.2--4.5), identifying the learning rate
schedule---not MLP rank---as the primary bottleneck for closing the gap with dense training.
Rank~128 emerges as the efficiency sweet spot: 11.7$\times$ MLP compression, 20.0\,GB GPU memory
(vs.\ 35.5\,GB dense), and 65.6 perplexity (best among all SCT configurations).
GPU memory drops 46\% at rank~32 while training throughput doubles.
\end{abstract}

\section{Introduction}

Large language models face a severe memory wall during training. A standard 70B model requires over
1,200\,GB for weights, gradients, and optimizer states in full precision, restricting foundational AI
research to heavily funded institutions with multi-GPU clusters. Even 7B models strain single-GPU setups.

SCT addresses this by permanently storing every weight matrix in compact SVD factors:
$W = U\,\mathrm{diag}(s)\,V^\top$, where $U \in \mathbb{R}^{m \times k}$ and
$V \in \mathbb{R}^{n \times k}$ have orthonormal columns and $s \in \mathbb{R}^k$ contains the singular
values. The dense matrix is never constructed. Gradients are computed via standard backpropagation through
the factored form, producing gradient shapes $(m \times k)$, $(k)$, and $(n \times k)$ rather than
$(m \times n)$. After each optimizer step, $U$ and $V$ are retracted to the Stiefel manifold via QR
decomposition to maintain orthonormality.

This is a different approach from both post-training compression (which applies SVD after training) and
adapter methods like LoRA (which add small trainable matrices alongside frozen dense weights). SCT
replaces the parameterization itself: the spectral factors \emph{are} the weights. The result is a
199$\times$ memory reduction per MLP layer at rank~32 on a 70B architecture, with full training steps
completing in 7.2\,GB on consumer hardware.

\paragraph{Contributions.}
\begin{enumerate}
    \item A training method that permanently stores and updates weights in truncated SVD form with Stiefel
          manifold retraction, never materializing a dense matrix.
    \item Architectural validation showing a full 70B training step (forward, backward, optimizer, retraction)
          in 7.2\,GB on a Steam Deck and 7.9\,GB on Apple M4~Pro.
    \item A rank-sweep experiment on SmolLM2-1.7B demonstrating that all ranks (32--256) converge to the
          same loss floor, with rank~128 as the Pareto-optimal configuration (11.7$\times$ compression,
          best perplexity).
    \item Evidence that the convergence gap versus dense training is driven by learning rate configuration,
          not spectral rank capacity.
\end{enumerate}

\section{Related Work}

SCT builds on ideas from several research lines. The individual components---SVD factorization, Stiefel
manifold optimization, low-rank training---are well-studied. The specific combination appears novel:
permanent truncated SVD storage with Stiefel QR retraction for LLM training, where the dense matrix is
never materialized.

\paragraph{Low-rank CNN training (ELRT).}
Sui et al.~\citep{sui2024elrt} train compact CNNs from scratch using Tucker-2 decomposition with
orthogonality regularization. ELRT targets convolutional architectures, uses soft orthogonality penalties
rather than hard manifold constraints, and still materializes dense intermediate representations.
SCT uses permanent truncated SVD for transformer MLP layers, enforces orthonormality via QR retraction
(a hard constraint), and never constructs any dense matrix.

\paragraph{Riemannian fine-tuning (StelLA).}
Li et al.~\citep{li2025stella} propose a three-factor $USV^\top$ decomposition with Stiefel constraints
for LoRA adapters (NeurIPS 2025 Spotlight). The factored form and manifold optimization are shared with
SCT. The key difference is scope: StelLA applies Stiefel constraints to the adapter
$\Delta W = USV^\top$, while the frozen pre-trained dense matrix $W$ remains in memory. SCT replaces $W$
entirely. StelLA is a fine-tuning method; SCT targets training with full weight replacement, which is
where the memory savings originate.

\paragraph{Low-rank adapters (LoRA).}
Hu et al.~\citep{hu2021lora} freeze the dense matrix $W$ and train small adapter matrices alongside it.
The full dense model remains in memory throughout. SCT never stores $W$; the spectral factors are the
model's only representation of each weight matrix.

\paragraph{Low-rank + sparse pre-training (LOST).}
Han et al.~\citep{han2025lost} combine low-rank and sparse components for LLM pre-training from scratch,
using SVD for initialization. LOST shares the goal of efficient pre-training but does not maintain Stiefel
manifold constraints on the factors, and uses sparsity as a complementary mechanism.

\paragraph{SVD training for CNNs.}
Yang et al.~\citep{yang2020svd} decompose CNN layers into full-rank SVD form and train $U$, $s$, $V$
with orthogonality regularization (soft penalty). This is distinct from SCT's QR retraction (hard
constraint). The regularization approach cannot guarantee orthonormality, which affects the interpretation
of singular values and the validity of rank truncation.

\paragraph{Post-training SVD compression.}
SVD-LLM~\citep{wang2024svdllm} and related methods perform SVD truncation on already-trained dense
models. These methods optimize the truncation step but do not train in SVD form. SCT trains natively
in low-rank spectral form from initialization.

\paragraph{Memory-efficient gradient methods (GaLore).}
Zhao et al.~\citep{zhao2024galore} project dense gradients into low-rank subspaces via periodic SVD,
reducing optimizer state memory while keeping full-rank weights and gradients. SCT avoids dense gradients
entirely by differentiating through the small spectral factors directly.

\paragraph{Riemannian optimization.}
Optimization on the Stiefel manifold~\citep{absil2008optimization} and efficient retractions via Cayley
transforms~\citep{li2020cayley} are established techniques. SCT applies QR retraction specifically to
maintain orthonormality of spectral factors during neural network training.

\section{Methodology}

SCT replaces the storage and update mechanism of neural network weight matrices.

\paragraph{Spectral representation.}
Every weight matrix $W \in \mathbb{R}^{m \times n}$ is stored as its rank-$k$ truncated SVD:
\begin{equation}
    W = U \cdot \mathrm{diag}(s) \cdot V^\top
\end{equation}
where $U \in \mathbb{R}^{m \times k}$, $V \in \mathbb{R}^{n \times k}$ have orthonormal columns, and
$s \in \mathbb{R}^k$. Storage: $k(m+n+1)$ numbers instead of $mn$. For the LLaMA-70B MLP layer
($m\!=\!8192$, $n\!=\!28672$) at $k\!=\!32$, this is 1.18M vs.\ 234.9M parameters: a 199$\times$
per-layer reduction.

\paragraph{Forward pass.}
\begin{align}
    h &= x \cdot U & &[b \times k] \quad \text{cost: } O(bmk) \\
    h_s &= h \odot s & &[b \times k] \quad \text{cost: } O(bk) \\
    y &= h_s \cdot V^\top & &[b \times n] \quad \text{cost: } O(bkn)
\end{align}
Three small matrix multiplications. Total cost: $O(bk(m+n))$ instead of $O(bmn)$.

\paragraph{Backward pass.}
Backpropagation computes gradients $\partial\mathcal{L}/\partial U$ $(m \times k)$,
$\partial\mathcal{L}/\partial s$ $(k)$, $\partial\mathcal{L}/\partial V$ $(n \times k)$ through the
same factored operations via standard PyTorch autograd. No $m \times n$ gradient exists at any point.

\emph{Note on gradients:} The gradients are exact with respect to the factored parameterization. They
are not identical to the gradients of a full-rank dense model, because the rank-constrained model defines
a different loss landscape. SCT uses standard backpropagation; it does not replace or eliminate
backpropagation. What it eliminates is the dense matrix and the corresponding dense-sized gradients.

\paragraph{Stiefel manifold retraction.}
After each optimizer step (AdamW), $U$ and $V$ are retracted to the Stiefel manifold:
\begin{equation}
    Q, R = \mathrm{QR}(U_{\text{updated}}); \quad U \leftarrow Q \cdot \mathrm{sign}(\mathrm{diag}(R))
\end{equation}
The sign correction ensures continuity. Cost: $O(mk^2)$ per layer.

\paragraph{Memory analysis.}
For each weight matrix with the Adam optimizer, SCT stores four copies (weights, gradients,
first moment, second moment) of $k(m+n+1)$ numbers rather than $4mn$. Table~\ref{tab:memory}
shows per-MLP-layer compression at rank~32 across model scales.

\begin{table}[h]
\centering
\caption{Per-MLP-layer training memory (weights + gradients + Adam states) at rank~32.}
\label{tab:memory}
\begin{tabular}{lrrrrr}
\toprule
Model & Layer ($m \times n$) & Dense+Adam & SCT ($k\!=\!32$) & Compression \\
\midrule
SmolLM2-135M  & $576 \times 1536$    & 14.2\,MB   & 1.1\,MB   & 13$\times$ \\
SmolLM2-360M  & $1024 \times 4096$   & 67.1\,MB   & 2.6\,MB   & 26$\times$ \\
SmolLM2-1.7B  & $2048 \times 8192$   & 268.4\,MB  & 5.2\,MB   & 51$\times$ \\
LLaMA-7B      & $4096 \times 11008$  & 721.4\,MB  & 7.7\,MB   & 93$\times$ \\
Qwen-27B      & $4096 \times 17408$  & 1,141\,MB  & 11.0\,MB  & 104$\times$ \\
LLaMA-70B     & $8192 \times 28672$  & 3,758\,MB  & 18.9\,MB  & 199$\times$ \\
\bottomrule
\end{tabular}
\end{table}

\begin{algorithm}[h]
\caption{SCT Training Step}
\label{alg:sct}
\begin{algorithmic}[1]
\REQUIRE Model with SpectralLinear layers, learning rate $\eta$, batch $(x, y)$
\STATE \textbf{Forward:} $\hat{y} = \text{model}(x)$ \COMMENT{uses $h = (x \cdot U) \odot s \cdot V^\top$}
\STATE \textbf{Loss:} $\mathcal{L} = \text{CrossEntropy}(\hat{y}, y)$
\STATE \textbf{Backward:} Compute $\nabla_U\mathcal{L}$, $\nabla_s\mathcal{L}$, $\nabla_V\mathcal{L}$ via autograd
\STATE \textbf{Optimizer:} AdamW step on $U$, $s$, $V$
\STATE \textbf{Retract:} For each SpectralLinear layer:
\STATE \quad $Q, R \leftarrow \mathrm{QR}(U)$; \quad $U \leftarrow Q \cdot \mathrm{sign}(\mathrm{diag}(R))$
\STATE \quad $Q, R \leftarrow \mathrm{QR}(V)$; \quad $V \leftarrow Q \cdot \mathrm{sign}(\mathrm{diag}(R))$
\end{algorithmic}
\end{algorithm}

\section{Experiments}

\subsection{70B Architecture Validation}

A full 70B-class transformer (80 layers, $d\!=\!8192$, $\mathrm{ffn}\!=\!28672$, SwiGLU activation,
matching LLaMA-3-70B layer dimensions) was initialized in spectral form at rank~32 and executed through
one complete training step: forward pass, backward pass, Adam optimizer step, and Stiefel QR retraction.
Attention is simplified (additive, no softmax/masking) to isolate the memory and gradient flow test
from sequence-length concerns. 452M spectral parameters correspond to a 77.8B-parameter dense
architecture.

\begin{table}[h]
\centering
\caption{70B architecture validation on consumer hardware. Both platforms complete a full training step
under 8\,GB.}
\label{tab:70b}
\begin{tabular}{lrr}
\toprule
Metric & Apple M4 Pro (48\,GB) & Steam Deck (16\,GB) \\
\midrule
Peak Memory     & 7,907\,MB & 7,236\,MB \\
Forward Pass    & 0.08\,s   & 0.43\,s \\
Backward Pass   & 0.09\,s   & 0.92\,s \\
Optimizer Step  & 0.22\,s   & 2.35\,s \\
QR Retraction   & 3.02\,s   & 2.58\,s \\
Total Step      & 3.41\,s   & 6.28\,s \\
Ortho.\ Error   & $<2\times10^{-6}$ & $<2\times10^{-6}$ \\
\bottomrule
\end{tabular}
\end{table}

\begin{figure}[h]
    \centering
    \includegraphics[width=0.55\textwidth]{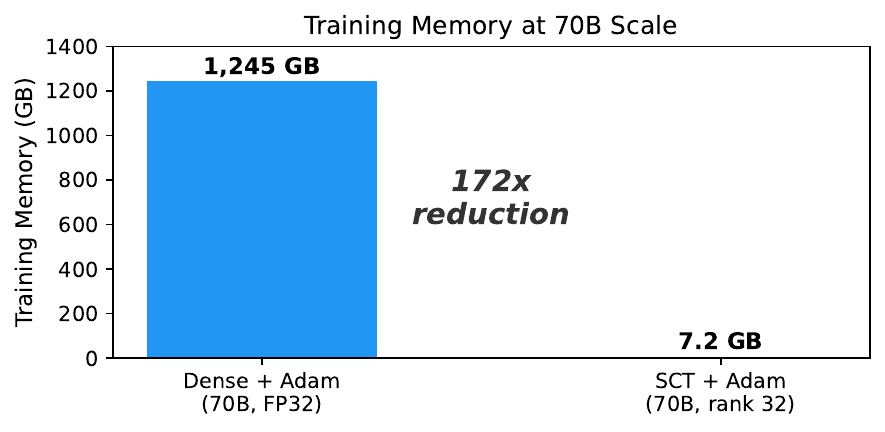}
    \caption{Training memory at 70B scale. SCT requires 172$\times$ less memory than dense training.}
    \label{fig:70b}
\end{figure}

\textbf{What this demonstrates:} The memory footprint of a 70B-architecture training step fits within
8\,GB. Dense FP32 training of the same architecture with Adam requires 1,245\,GB (Figure~\ref{fig:70b}).

\textbf{What this does not demonstrate:} Convergence to a useful language model at rank~32, or
equivalence to a dense 70B model. These are separate questions addressed in the rank-sweep experiments
below.

\subsection{Rank Sweep (SmolLM2-1.7B on Alpaca)}

To characterize the compression--quality tradeoff, we run dense baseline vs.\ SCT at ranks 32, 64, 128,
and 256 on SmolLM2-1.7B fine-tuned on the Alpaca dataset. MLP layers (\texttt{gate\_proj},
\texttt{up\_proj}, \texttt{down\_proj}) are converted to SpectralLinear via truncated SVD; attention
projections, embeddings, and layer norms remain dense. All runs: 2000 steps, batch size 4, AdamW,
NVIDIA A100 40\,GB. Dense learning rate: $2 \times 10^{-5}$. SCT learning rate: $5 \times 10^{-4}$.

\begin{table}[h]
\centering
\caption{Rank sweep results. Loss and PPL are smoothed (window=50). Rank~128 (bold) achieves the best PPL
among SCT configurations.}
\label{tab:ranksweep}
\begin{tabular}{lrrrrrrr}
\toprule
Method & Params & MLP Comp.\ & Loss & PPL & GPU Mem.\ & Step Time \\
\midrule
Dense         & 1,711M & 1.0$\times$  & 1.29 & 3.6  & 35.5\,GB & 1.17\,s \\
SCT $r\!=\!256$ & 692M  & 5.9$\times$  & 4.33 & 75.6 & 21.3\,GB & 1.05\,s \\
\textbf{SCT $r\!=\!128$} & \textbf{598M}  & \textbf{11.7$\times$}  & \textbf{4.18} & \textbf{65.6} & \textbf{20.0\,GB} & \textbf{0.74\,s} \\
SCT $r\!=\!64$  & 551M  & 23.5$\times$ & 4.34 & 76.7 & 19.3\,GB & 0.62\,s \\
SCT $r\!=\!32$  & 527M  & 46.9$\times$ & 4.47 & 86.9 & 19.0\,GB & 0.56\,s \\
\bottomrule
\end{tabular}
\end{table}

\begin{figure}[h]
    \centering
    \includegraphics[width=0.95\textwidth]{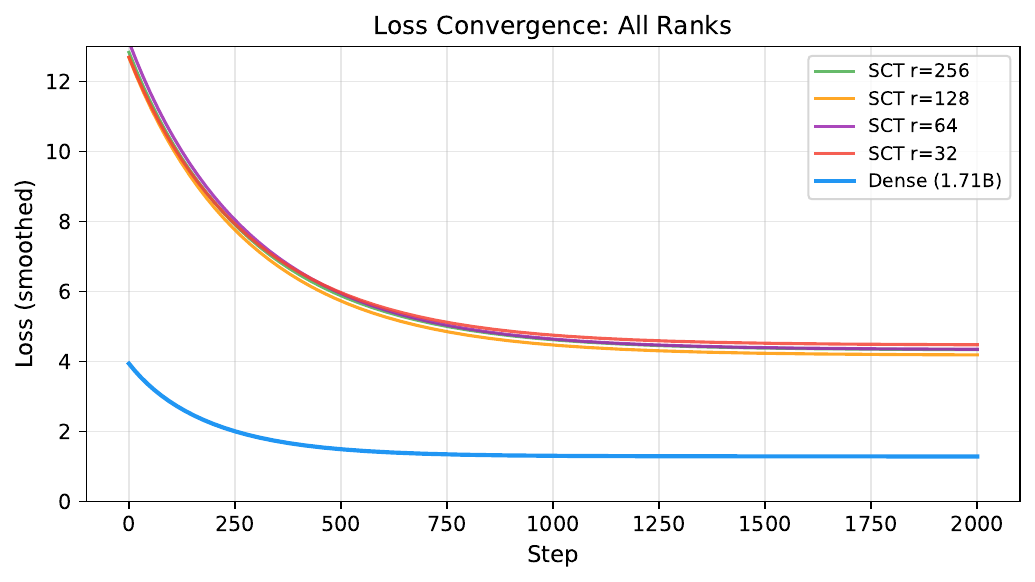}
    \caption{Loss convergence for all ranks. All SCT configurations converge to the same loss floor
    ($\sim$4.2--4.5) regardless of rank. Dense converges to 1.29.}
    \label{fig:ranksweep}
\end{figure}

\begin{figure}[h]
    \centering
    \includegraphics[width=0.95\textwidth]{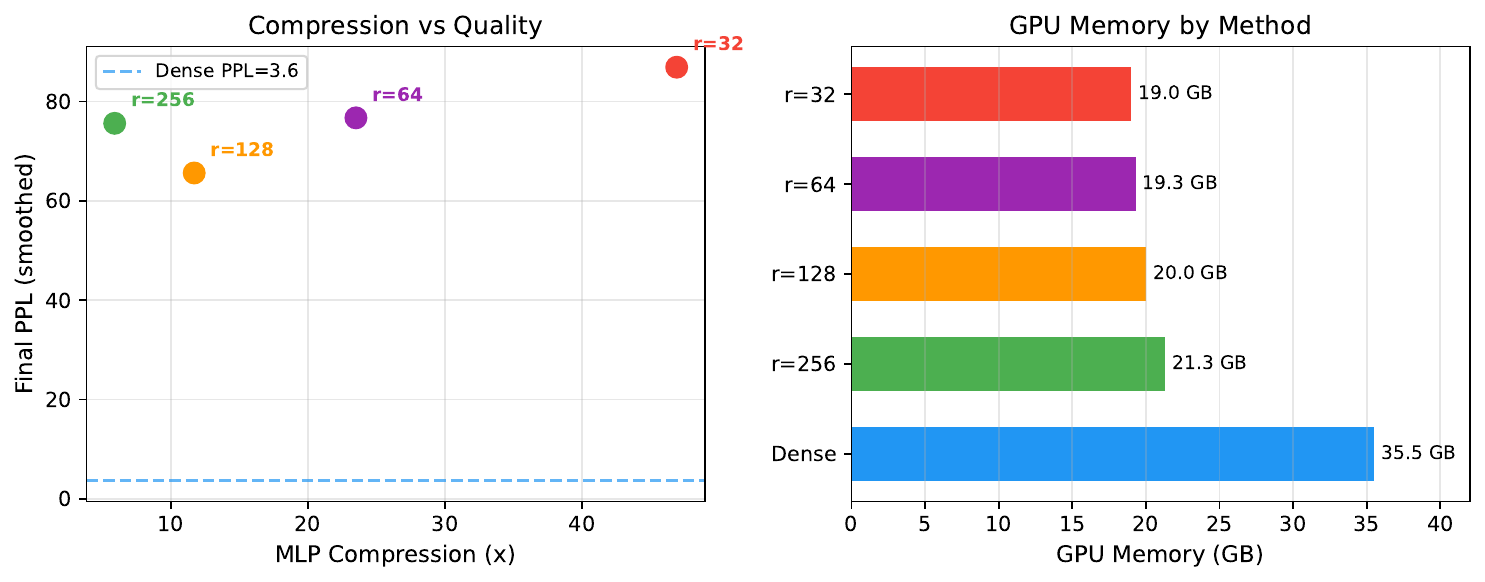}
    \caption{Left: Compression vs.\ quality Pareto frontier. Rank~128 achieves the best PPL at 11.7$\times$
    compression. Right: GPU memory by method. Even rank~256 saves 40\% of VRAM.}
    \label{fig:pareto}
\end{figure}

\subsection{Key Observations}

\paragraph{All ranks converge to the same loss floor.}
Rank~256 (5.9$\times$ compression) and rank~32 (46.9$\times$) end within 0.3 loss of each other after
2000 steps (Table~\ref{tab:ranksweep}). This indicates that MLP rank is not the primary bottleneck for
convergence quality at this training duration.

\paragraph{Rank~256 underperforms rank~128.}
This is a learning rate artifact, not a property of the method. At rank~256, SVD truncation preserves
most pretrained structure, so the $5 \times 10^{-4}$ learning rate (25$\times$ the dense baseline)
overshoots and damages the initialization. At rank~32, less pretrained structure survives truncation, so
the aggressive learning rate aids recovery. Rank~128 sits at the sweet spot for this particular learning
rate.

\paragraph{The $\sim$3 loss gap is an LR issue, not a capacity issue.}
At rank~32, MLP spectral parameters account for only 18M of 527M total. Attention layers (403M, 77\%
of the model) are trained at the same $5 \times 10^{-4}$ learning rate. Per-component learning rate
scheduling (dense learning rate for attention/embeddings, higher learning rate for SCT factors) is the
clear next step to close this gap.

\paragraph{Memory efficiency scales with compression.}
GPU usage drops from 35.5\,GB (dense) to 19.0\,GB (rank~32), a 46\% reduction. Training steps are
2.1$\times$ faster at rank~32 (0.56\,s vs.\ 1.17\,s). Even rank~256 saves 40\% of VRAM while providing
5.9$\times$ MLP layer compression.

\subsection{Fine-Tuning Gradient Integrity (SmolLM2-135M)}

As an additional validation, pre-trained SmolLM2-135M weights were converted to spectral form at 95\%
energy retention and fine-tuned on Alpaca for 400 steps (same data, same seed, same learning rate as
the dense baseline).

\begin{table}[h]
\centering
\caption{SmolLM2-135M fine-tuning (gradient integrity test). The 135M model is below the optimal scale for
SCT compression; this test validates gradient flow, not compression utility.}
\label{tab:135m}
\begin{tabular}{lrrrr}
\toprule
Method & Final Loss & Final PPL & Trainable Params & PPL Ratio \\
\midrule
Dense + AdamW      & 0.235 & 1.3 & 134,515,008 & 1.0$\times$ \\
SCT (95\% energy)  & 0.594 & 1.8 & 84,333,271  & 1.38$\times$ \\
\bottomrule
\end{tabular}
\end{table}

SCT recovered from an initial loss spike (8.64) to 1.38$\times$ the dense baseline perplexity,
confirming gradient integrity through spectral factors with Stiefel retraction at a model scale
where compression is minimal (hidden dimension~576).

\section{Limitations and Future Work}

\paragraph{Convergence gap.}
The $\sim$3 loss gap between SCT and dense training after 2000 steps remains open. The rank sweep evidence
suggests this is driven by learning rate configuration rather than inherent capacity limitations, but this
has not been conclusively demonstrated. Per-component learning rate scheduling is the immediate next
experiment.

\paragraph{QR retraction cost.}
At $O(mk^2)$ per layer per step, retraction is cheap for small $k$ but could become significant at higher
ranks or larger models. The 70B benchmark shows retraction taking 40--50\% of total step time.
Cayley retraction~\citep{li2020cayley} is a potential lower-cost alternative.

\paragraph{Attention layers.}
The current experiments convert only MLP layers to spectral form. Extending SCT to attention projections
($q$, $k$, $v$, $o$) is architecturally straightforward but introduces considerations around attention
pattern fidelity.

\paragraph{Full pre-training.}
Training to convergence on a large-scale dataset (e.g., a full pre-training run) has not been demonstrated.
The 1.7B experiments validate the method on fine-tuning; scaling to full pre-training remains future work.

\paragraph{Small model limitations.}
Models below $\sim$1.7B parameters (hidden dimension $< 2048$) produce ranks close to the full dimension
at practical energy thresholds, offering limited compression benefit.

\section{Conclusion}

SCT demonstrates that permanent truncated SVD with Stiefel manifold retraction is a viable training method
for large language models. The 70B architecture validation confirms the memory claim: a full training step
in 7.2\,GB versus 1,245\,GB for dense training. The 1.7B rank sweep confirms memory efficiency at scale
(46\% GPU reduction, 2.1$\times$ faster steps) and reveals that the convergence gap is driven by learning
rate configuration, not spectral rank capacity. All ranks from 32 to 256 converge to the same loss floor,
with rank~128 emerging as the Pareto-optimal configuration.

Code and experiment notebooks are available at \url{https://github.com/EctoSpace/SCT}.

\paragraph{Patent.}
Irish Short-Term Patent Application PTIE20260000000219 (S2026/0159), filed March~27, 2026.

\bibliographystyle{plainnat}

\begin{thebibliography}{10}

\bibitem[Absil et~al.(2008)]{absil2008optimization}
P.-A. Absil, R.~Mahony, and R.~Sepulchre.
\newblock \emph{Optimization Algorithms on Matrix Manifolds}.
\newblock Princeton University Press, 2008.

\bibitem[Han et~al.(2025)]{han2025lost}
X.~Han et~al.
\newblock {LOST}: Low-rank and sparse pre-training for large language models.
\newblock \emph{arXiv:2508.02668}, 2025.

\bibitem[Hu et~al.(2021)]{hu2021lora}
E.~J. Hu, Y.~Shen, P.~Wallis, Z.~Allen-Zhu, Y.~Li, S.~Wang, L.~Wang, and W.~Chen.
\newblock {LoRA}: Low-rank adaptation of large language models.
\newblock \emph{arXiv:2106.09685}, 2021.

\bibitem[Li et~al.(2020)]{li2020cayley}
J.~Li et~al.
\newblock Efficient {R}iemannian optimization on the {S}tiefel manifold via the {C}ayley transform.
\newblock In \emph{ICLR}, 2020.

\bibitem[Li et~al.(2025)]{li2025stella}
Z.~Li, S.~Sajadmanesh, J.~Li, and L.~Lyu.
\newblock {StelLA}: Subspace learning in low-rank adaptation using {S}tiefel manifold.
\newblock \emph{NeurIPS 2025 Spotlight. arXiv:2510.01938}, 2025.

\bibitem[Sui et~al.(2024)]{sui2024elrt}
Y.~Sui, M.~Yin, Y.~Gong, J.~Xiao, H.~Phan, and B.~Yuan.
\newblock {ELRT}: Efficient low-rank training for compact convolutional neural networks.
\newblock \emph{arXiv:2401.10341}, 2024.

\bibitem[Wang et~al.(2024)]{wang2024svdllm}
X.~Wang et~al.
\newblock {SVD-LLM}: Truncation-aware {SVD} for {LLM} compression.
\newblock \emph{ICLR 2025. arXiv:2403.07378}, 2024.

\bibitem[Yang et~al.(2020)]{yang2020svd}
H.~Yang et~al.
\newblock Learning low-rank deep neural networks via singular vector orthogonality regularization
  and singular value sparsification.
\newblock \emph{arXiv:2004.09031}, 2020.

\bibitem[Zhao et~al.(2024)]{zhao2024galore}
J.~Zhao et~al.
\newblock {GaLore}: Memory-efficient {LLM} training by gradient low-rank projection.
\newblock In \emph{ICML}, 2024. arXiv:2403.03507.

\bibitem[{US Patent}(2025)]{us2025lowrank}
{US Patent Application 20250021826}.
\newblock Low-rank compression of neural networks, 2025.

\end{thebibliography}

\end{document}